\documentclass[acmsmall]{acmart}
\AtBeginDocument{%
  \providecommand\BibTeX{{%
    \normalfont B\kern-0.5em{\scshape i\kern-0.25em b}\kern-0.8em\TeX}}}

\setcopyright{acmlicensed}
\acmJournal{TIST}
\acmYear{2022} \acmVolume{1} \acmNumber{1} \acmArticle{1} \acmMonth{1} \acmPrice{15.00}\acmDOI{10.1145/3510031}

\acmJournal{JACM}

\usepackage{xcolor}
\usepackage{times}
\usepackage{helvet}
\usepackage{courier}
\usepackage{url}
\usepackage{amsmath}
\usepackage{amsthm}
\usepackage{graphicx}
\usepackage{multicol}
\usepackage{algorithm}
\usepackage{algorithmic}
\usepackage{braket}
\usepackage{array}
\usepackage{tabu}
\usepackage{subfigure}
\usepackage{epstopdf}
\begin{document}

\title{FedCVT: Semi-Supervised Vertical Federated Learning with Cross-View Training}



\author{Yan Kang}
\email{yangkang@webank.com}
\affiliation{%
  \institution{WeBank}
  \city{Shenzhen}
  \country{China}}

\author{Yang Liu}
\email{liuy03@air.tsinghua.edu.cn}
\authornote{corresponding author}
\affiliation{%
  \institution{Institute for AI Industry Research, Tsinghua University}
  \city{Beijing}
  \country{China}
}

\author{Xinle Liang}
\email{shindler@mail.ustc.edu.cn}
\affiliation{%
  \institution{WeBank}
  \city{Shenzhen}
  \country{China}
}


\renewcommand{\shortauthors}{Kang, et al.}

\begin{abstract}
Federated learning allows multiple parties to build machine learning models collaboratively without exposing data. In particular, vertical federated learning (VFL) enables participating parties to build a joint machine learning model based upon distributed features of aligned samples. However, VFL requires all parties to share a sufficient amount of aligned samples. In reality, the set of aligned samples may be small, leaving the majority of the non-aligned data unused. In this paper, we propose Federated Cross-View Training (FedCVT), a semi-supervised learning approach that improves the performance of the VFL model with limited aligned samples. More specifically, FedCVT estimates representations for missing features, predicts pseudo-labels for unlabeled samples to expand the training set, and trains three classifiers jointly based upon different views of the expanded training set to improve the VFL model's performance. FedCVT does not require parties to share their original data and model parameters, thus preserving data privacy. We conduct experiments on NUS-WIDE, Vehicle, and CIFAR10 datasets. The experimental results demonstrate that FedCVT significantly outperforms vanilla VFL that only utilizes aligned samples. Finally, we perform ablation studies to investigate the contribution of each component of FedCVT to the performance of FedCVT.

\end{abstract}



\maketitle

\section{Introduction}

With the increasingly stricter privacy-protection laws implemented worldwide \cite{GDPR, DLAPiper2019}, federated learning has received significant attention and become a popular research topic recently \cite{kairouz2019advances}. As the research goes deeper and wider, the practice of federated learning has been expanded from building powerful mobile applications based on data resided in millions of mobile devices \cite{DBLP:journals/corr/McMahanMRA16} to solving the problem of data silos among or within organizations \cite{flbook}. For example, many business decisions of a bank may rely on its customers' purchasing preferences. This bank may share some customers with a local retail company that owns local people's purchasing preference data. Thus, the bank can invite the retail company to build a joint model collaboratively by leveraging the data features of both sides to improve its business. The retail company can also benefit from this joint model.

This and various other similar practical demands \cite{flbook} motivate the development of \textit{vertical federated learning (VFL)}\cite{Yang-et-al:2019} (also called feature-partitioned federated learning), which enables participating parties to train a joint machine learning model collaboratively by utilizing scattered features of their aligned samples without disclosing original data. However, a critical prerequisite of VFL is that it requires all parties to share a sufficient amount of aligned samples to achieve competitive model performance. \cite{SecureFTL} proposes a federated transfer learning framework to address weak supervision (few labels) problems in the VFL setting. Nonetheless, it does not take full advantage of unlabeled samples for improving learning quality, and it aims to build a model only for the party of the target domain. Some other similar approaches, such as domain adaptation \cite{peng2019federated,peterson2019private,Liu_2021_CVPR} and knowledge distillation \cite{li2019fedmd,Gong_2021_ICCV}, have been applied in the federated learning setting. However, they mainly focus on scenarios where all parties share the same feature space (also known as \textit{horizontal federated learning} \cite{Yang-et-al:2019}). Therefore, research on applying semi-supervised techniques in the VFL setting is insufficient.

This paper proposes a novel semi-supervised algorithm in the VFL setting, termed FedCVT, to address the limitations of existing vertical federated learning approaches. Our contributions are as follows:

\begin{enumerate}
    \item FedCVT significantly boosts up the performance of the federated model when the aligned samples between participating parties are limited;
    \item FedCVT works with data of various types. This feature is important in practice since real-world VFL applications often need to deal with heterogeneous features;
    \item FedCVT does not require participating parties to share their original data and model parameters, but only intermediate representations and gradients, which can be further protected by the privacy-preserving VFL-DNN (Deep Neural Network) framework proposed by~\cite{yankang2021prada} and implemented in FATE~\cite{fate2021JMLR}\footnote{\url{https://github.com/FederatedAI/FATE}}.
\end{enumerate}

\section{Related Work}

In this section, we focus on reviewing closely related fields and approaches. 
\paragraph{Vertical Federated Learning (VFL)} VFL is also known as feature-partitioned machine learning in some literature. The secure linear machine learning with data partitioned in the feature space has been well studied in \cite{Gascn2016SecureLR,Hardy2017PrivateFL,DBLP:journals/iacr/MohasselZ17}. They apply either hybrid secure multi-party computation (SMPC) protocol or homomorphic encryption (HE) \cite{Rivest1978} to protect privacy in model training and inference. \cite{SecureBoost} proposes a secure federated tree-boosting (SecureBoost) approach in the VFL setting. It enables participating parties with different features to build a set of boosting trees collaboratively and proves that the SecureBoost provides the same level of accuracy as its non-privacy-preserving centralized counterparts. FATE~\cite{fate2021JMLR} designed and implemented a VFL-DNN (Deep Neural Network) framework that supports DNN in the VFL setting. VFL-DNN leverages a hybrid encryption scheme in the forward and backward stages of the VFL training procedure to protect data privacy and model security.
\cite{DBLP:journals/corr/abs-1812-00564} proposed several configurations of splitting a deep neural network to support various forms of collaborations among health facilities that each holds a partial deep network and owns a different modality of patient data. 

\paragraph{Semi-Supervised Learning (SSL)} 

SSL aims to alleviate the need for numerous labeled data to train machine learning models by leveraging unlabeled data. Recent semi-supervised learning works utilize transfer learning \cite{Pan:2010:CSC:1772690.1772767}, consistency regularization \cite{NIPS2017_6719,tm_ssl_2017} and pseudo-labeling \cite{pseudolbl_icml2013,clark-etal-2018-semi} to learn from unlabeled data. \cite{DBLP:journals/corr/abs-1905-02249} proposed MixMatch that unifies three dominant semi-supervised methods: entropy minimization \cite{10.5555/2976040.2976107}, consistency regularization and generic regularization. It demonstrates that MixMatch can approach similar error rates as fully supervised training but depends on significantly fewer training data. \cite{clark-etal-2018-semi} proposed a semi-supervised Cross-View Training (CVT) approach that simultaneously trains a full model based upon all labeled data and multiple auxiliary models that only see restricted views of the unlabeled data. During training, CVT teaches auxiliary models to match predictions made by the full model to improve representation learning, thereby improving the performance of the full model. While successful, these works are not designed for VFL scenarios where features are scattered among parties. \cite{SecureFTL} proposed a secure Federated Transfer Learning (FTL) framework, which is the first framework that enables VFL to benefit from transfer learning. FTL helps the target domain party build a competitive prediction model by utilizing rich label resources from the source domain.

\section{Problem Definition}


We consider a two-party vertical federated learning setting where only one party has labels. This is the typical VFL setting defined in~\cite{flbook} where the party having labels is short of features to build an accurate model, and thus it leverages complimentary features provided by a second party. Specifically, party A has dataset $\mathcal{D}^A := \{(X^{A,i},Y^{A,i})\}_{i=1}^{n^A}$
where  $X^{A,i}$ is the feature vector of the $i \text{th}$ sample and $Y^{A,i} \in \{0,1\}^C$ is the corresponding one-hot encoding ground-truth label with $C$ classes for the $i$th sample, while party B has dataset $\mathcal{D}^B :=\{X^{B,i}\}_{i=1}^{n^B}$. $\mathcal{D}^A$ and $\mathcal{D}^B$ are held privately by the two parties and cannot be exposed to each other. $n^{A}$ and $n^B$ are numbers of samples for $\mathcal{D}^A$ and $\mathcal{D}^B$, respectively. 



\begin{figure}[hb]
\centering
\includegraphics[width=0.70\linewidth]{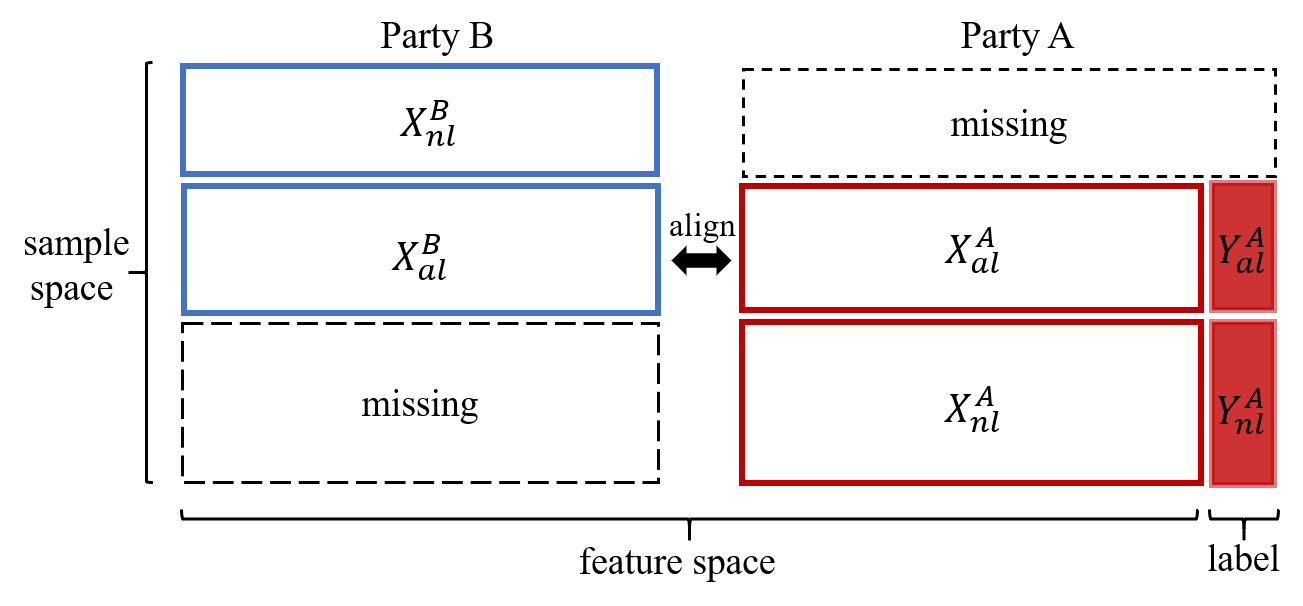}
\caption{View of the virtual dataset in vertical federated learning. Each party owns a vertical partition (or a partial view) of this dataset.} \label{fig_vfl}
\end{figure}

If we concatenate $\mathcal{D}^{A}$ and $\mathcal{D}^{B}$ with samples (of different parties) having the same identity aligned, we obtain a single dataset depicted in Figure \ref{fig_vfl}. This dataset is vertically partitioned, and each party owns one vertical partition (or a partial view) of this dataset. This is where the term "vertical federated learning" comes from. We assume that there exists only a limited amount of aligned samples $\mathcal{D}_{al} := \{X_{al}^{B,i},X_{al}^{A,i},Y_{al}^{A,i}\}_{i=1}^{n_{al}}$ between the two parties. Party A owns the partition $\mathcal{D}^A_{al} := \{X_{al}^{A,i},Y_{al}^{A,i}\}_{i=1}^{n_{al}}$ and party B owns $\mathcal{D}^B_{al} := \{X_{al}^{B,i}\}_{i=1}^{n_{al}}$, where $n_{al}$ is the number of aligned samples. One can perform the alignment through encrypted entity matching techniques in a privacy-preserving setting \cite{er_nock_2018}. Here, we assume that party A and party B have already established the alignment between their samples. 

Other than aligned samples, the rest samples of each party have no aligned counterparts from the other party. We call them non-aligned samples and denote them as $\mathcal{D}^A_{nl}:=\{X_{nl}^{A, i},Y_{nl}^{A, i}\}_{i=1}^{n^A_{nl}}$ for party A  and $\mathcal{D}^B_{nl}:=\{X_{nl}^{B, i}\}_{i=1}^{n^B_{nl}}$ for party B. From the perspective of a single tabular dataset (see Figure \ref{fig_vfl}), each party has no features (or labels) that correspond to non-aligned samples of the other party. We treat those features (or labels) are \emph{missing}.




The conventional VFL is trying to build a federated machine learning model utilizing only aligned samples $D_{al}$, leaving non-aligned samples $D^A_{nl}$ and $D^B_{nl}$ unused. We, therefore, propose a semi-supervised VFL with Cross-View Training (FedCVT) approach that not only leverages aligned samples but also takes full advantages of non-aligned samples (shown in Figure \ref{overview}), aiming to boost the performance of VFL, especially when the amount of aligned samples is small.


\section{Overview}

\begin{figure}[ht]
\centering
\includegraphics[width=0.98\linewidth]{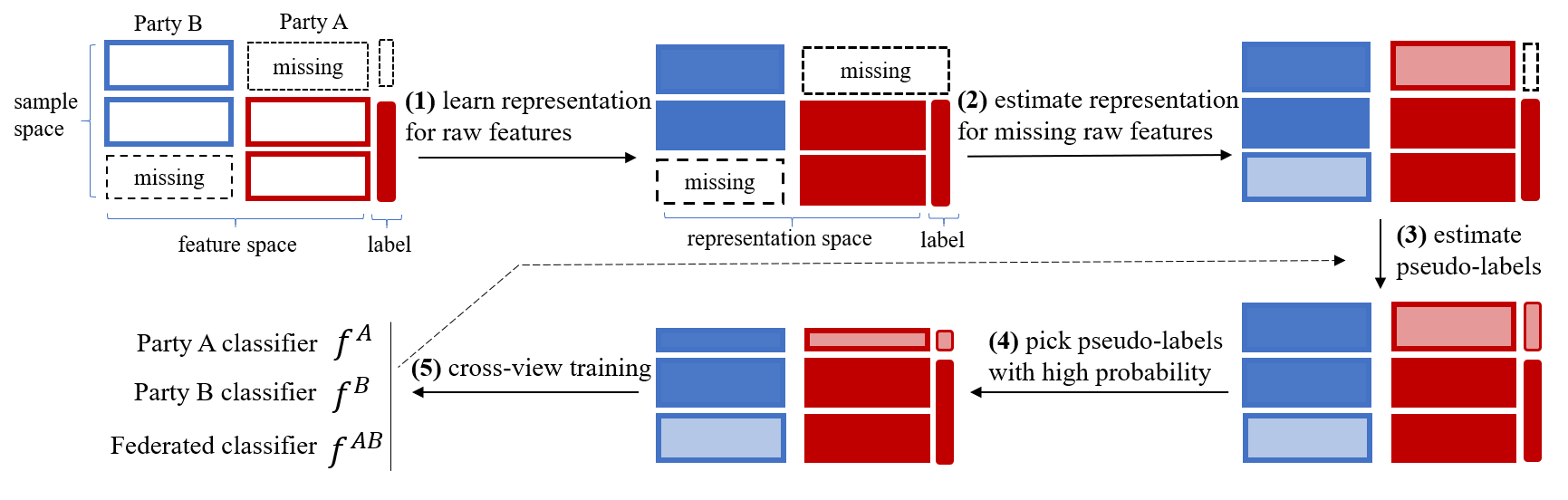}
\caption{Overview of FedCVT approach}\label{overview}
\end{figure}

Figure \ref{overview} depicts the overall workflow of our proposed FedCVT approach, which involves five stages:
\begin{enumerate}
    \item Learn representations from raw input features using neural networks;
    \item Estimate representations for missing features based on learned representations;
    \item Apply three trained classifiers $f^A$, $f^B$, and $f^{AB}$ to predict three candidate pseudo-labels for each unlabeled sample;
    \item When all the three candidate pseudo-labels for a sample are equal, and their probabilities are above a predefined threshold, retain this pseudo-labeled sample in the training set;
    \item Train the three classifiers $f^A$, $f^B$, and $f^{AB}$ jointly through vertical federated learning that each classifier takes as input a different view of the training dataset.
\end{enumerate}

\section{The Proposed Approach}

As depicted in Figure \ref{overview}, our FedCVT approach involves the following stages: learn representations from raw features (section \ref{repr_learning}); estimate representations for missing features (section \ref{est_repr_sec}); estimate pseudo-labels for unlabeled samples and cherry-pick pseudo-labeled samples (section \ref{est_lbl}); finally, jointly train three classifiers through vertical federated learning (section \ref{mvt}). In this section, we describe each of these stages in detail.

\subsection{Representation Learning} \label{repr_learning}

Deep neural networks have been widely used to learn representations \cite{Oquab14}. In this work, we utilize two neural networks for each party to learn representations from raw input data. One is to learn representations weakly shared between the two parties and the other is to learn representations unique to each party. 

\begin{figure}[ht]
\centering
\includegraphics[width=0.65\linewidth]{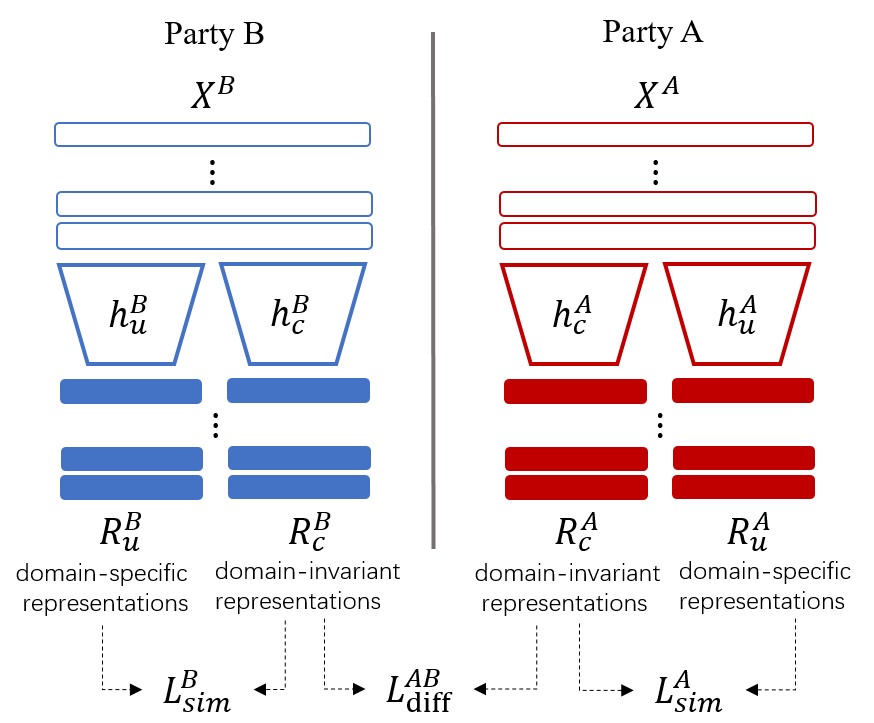}
\caption{Learn representations from raw input features. Each party $p \in \{A, B\}$ has two neural network models $\it{h}_u^p$ and $\it{h}_c^p$ to learn unique and shared representations, respectively.}\label{nn}
\end{figure}

As shown in Figure \ref{nn}, for each party $p \in \{A, B\}$, we denote $R_u^{p}=\it{h}_u^p(X^p)$ as the \emph{unique} representations and $R_c^{p}=\it{h}_c^p(X^p)$ as the \emph{shared} representations that are learned from raw input $X^p$ through neural networks $\it{h}_u^p$ and $\it{h}_c^p$, respectively. $R_u^p\in \mathbb{R}^{N^p\times d^p}$ and $R_c^p\in \mathbb{R}^{N^p\times d^p}$, where $\it{d^p}$ is the dimension of the top hidden representation layer of neural networks of party $p$. The complete learned representations for $X^p$ is denoted as $R^{p}= [R_c^p;R_u^p]$, where $[;]$ denotes the concatenation operator that concatenates two matrices along the feature axis. Further, we denote $R_{al}^{p}= [R_{c,al}^p;R_{u,al}^p]$ as representations learned from aligned samples $X_{al}^p$, while $R_{nl}^{p}= [R_{c,nl}^p;R_{u,nl}^p]$ as representations learned from non-aligned samples $X_{nl}^p$.

Intuitively, neural network $\it{h}_c^p$ aims to capture domain-invariant representations between the two parties while $\it{h}_u^p$ helps learn domain-specific representations. We propose following three loss terms to enforce neural networks to learn the desired representations.

\begin{equation}\label{shared_reprs_sim}
L^{AB}_{\text{diff}}(R^{A}_{c, al}, R^{B}_{c, al}) = \frac{1}{n_{al}}\sum_{i}\left\| R^{A,i}_{c, al}- R^{B,i}_{c, al}\right\|^2_F
\end{equation}
\begin{equation}\label{A_reprs_dist}
L^A_{sim}(R^{A}_{u}, R^{A}_{c}) = \frac{1}{n^{A}}\sum_{i}\left\| R^{A,i}_{u} \otimes R^{A,i}_{c}\right\|^2_F
\end{equation}
\begin{equation}\label{B_reprs_dist}
L^B_{sim}(R^{B}_{u}, R^{B}_{c}) = \frac{1}{n^{B}}\sum_{i}\left\| R^{B,i}_{u} \otimes R^{B,i}_{c}\right\|^2_F
\end{equation}

where $\otimes$ denotes matrix multiplication operator and $\left\| \cdot \right\|^2_F$ denotes the squared Frobenius norm. By minimizing (\ref{shared_reprs_sim}), neural networks $h_c^A$ and $h_c^B$ are pushed to learn shared representations from the raw features of the two parties. While (\ref{A_reprs_dist}) and (\ref{B_reprs_dist}) are orthogonality constraints \cite{NIPS2016_6254} that encourage neural networks $h_c^p$ and $h_u^p$ belonging to party $p \in \{A, B\}$ to learn distinct representations.

\subsection{Representation Estimation} 
\label{est_repr_sec}


Instead of directly imputing values of missing features, we estimate representations of \textit{missing} features (see Figure \ref{fig_vfl}). In other words, we want to estimate missing representations of party A that correspond to non-aligned representations $R_{nl}^{B}$ of party B and estimate missing representations of party B that correspond to non-aligned representations $R_{nl}^{A}$ of party A.

Each missing representation is estimated through the weighted sum over representations learned from raw features. The weights reflect the similarity between the representation to be estimated and the learned representations. To compute these weights, we leverage the scaled dot-product attention (SDPA) function \cite{NIPS2017_7181}:
\begin{equation}\label{sdpa}
   g(Q, K, V) = \text{softmax}(\frac{Q \otimes K^T}{\sqrt{d_k}}) \otimes V
\end{equation}

where $Q$ is the query matrix, $K$ is the key matrix, $V$ is the value matrix, and $Q$ and $K$ have the same dimension denoted as $d_k$. 

In the rest of this section, we will elaborate on how the SDPA function $g$ is applied to performing the representation estimation by giving an example of estimating party A's missing representations, denoted as  $\tilde{R}^{A}$, that correspond to the representations $R^{B}_{nl}$ learned from the non-aligned samples $X_{nl}^{B}$ of party B. Because $\tilde{R}^{A}=[\tilde{R}^{A}_c;\tilde{R}^{A}_u]$, the estimation of $\tilde{R}^{A}$ involves two parts: one is computing $\tilde{R}^{A}_c$ corresponding to shared representation $R^{B}_{nl,c}$ (section \ref{s_repr_est}) and another is computing $\tilde{R}^{A}_u$ corresponding to unique representation $R^{B}_{nl,u}$ (section \ref{u_repr_est}).


\begin{figure}[ht]
\centering
\includegraphics[width=0.95\linewidth]{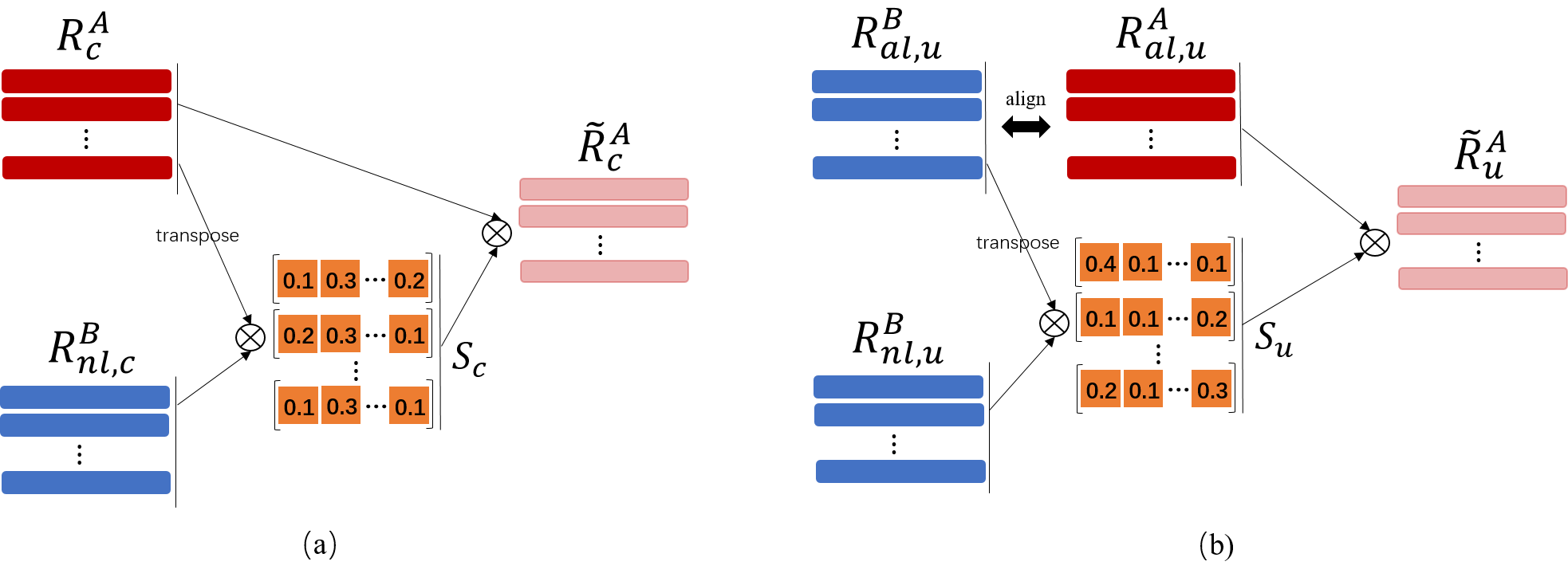}
\caption{Estimate missing representations $\tilde{R}^{A}$ corresponds to the non-aligned representations $R^{B}_{nl}$ of party B. (a) estimates missing representations $\tilde{R}^{A}_{c}$ corresponding to shared representations $R^{B}_{nl,c}$ while (b) estimates missing representations $\tilde{R}^{A}_{u}$ corresponding to unique representations $R^{B}_{nl,u}$. Then, $\tilde{R}^{A}$ can be obtained by concatenating $\tilde{R}^{A}_{c}$ and $\tilde{R}^{A}_{u}$.}
\label{est_repr}
\end{figure}

\subsubsection{Shared Representation Estimation}
\label{s_repr_est}

Formula (\ref{rAc}) estimates the \emph{shared} representation $\tilde{R}^{A}_{c}$ corresponding to $R^{B}_{nl,c}$:

\begin{equation}\label{rAc}
    \tilde{R}^{A}_{c} = g(R^{B}_{nl,c}, R^A_{c}, R^A_{c}) = \text{softmax}(\frac{R^{B}_{nl,c} \otimes (R^A_{c})^T}{\sqrt{d}}) \otimes R^A_{c}
\end{equation}

where $R_c^A$ denotes the shared representations of $X^A$ and $d$ denotes the dimension of $R_c^A$. The estimation procedure can be divided into the following two steps and it is pictorially described in Figure \ref{est_repr}(a).

\begin{itemize}
    \item \textbf{Step 1}: compute similarities between each representation vector of $R^{B}_{nl,c}$ and that of $R^{A}_c$. The result is a similarity matrix, denoted as $S_c$;
    \item \textbf{Step 2}: compute $\tilde{R}^{A}_c$ by $\text{softmax}(\frac{S_c}{\sqrt{d}}) \otimes R^{A}_c$.
\end{itemize}


\subsubsection{Unique Representation Estimation} 
\label{u_repr_est}

Formula (\ref{rAu}) estimates the \emph{unique} representation $\tilde{R}^{A}_{u}$ corresponding to $R^{B}_{nl,u}$.

\begin{equation}\label{rAu}
\tilde{R}^{A}_{u} = g(R^{B}_{nl,u}, R^B_{al,u},R^A_{al,u}) = \text{softmax}(\frac{R^{B}_{nl,u} \otimes (R^B_{al,u})^T}{\sqrt{d}}) \otimes R^A_{al,u}
\end{equation}

where $R_{al,u}^A$ and $R_{al,u}^B$ denotes the unique representations of $X^A_{al}$ and $X^B_{al}$ respectively, and $d$ denotes the dimension of $R_{al,u}^B$. The estimation procedure can be divided into the following two steps and it is pictorially described in Figure \ref{est_repr}(b).

\begin{itemize}
    \item \textbf{Step 1}: compute similarities between each representation vector of $R^{B}_{nl,u}$ and that of $R^{B}_{al, u}$. This result is a similarity matrix, denoted as $S_u$;
    \item \textbf{Step 2}: compute $\tilde{R}^{A}_u$ by $\text{softmax}(\frac{S_u}{\sqrt{d}}) \otimes R^{A}_{al,u}$. 
\end{itemize}

Note that the similarity matrix $S_u$ is calculated based on $R_{al,u}^B$ in the representation space of party B and then applied to $R_{al, u}^A$ in the representation space of party A. We assume $S_u$ calculated based on $R_{al, u}^B$ can be applied to $R_{al, u}^A$ because $R_{al,u}^B$ and $R_{al,u}^A$ are aligned indicating that they are different views of the same samples.

\begin{figure}[hb]
\centering
\includegraphics[width=1.0\linewidth]{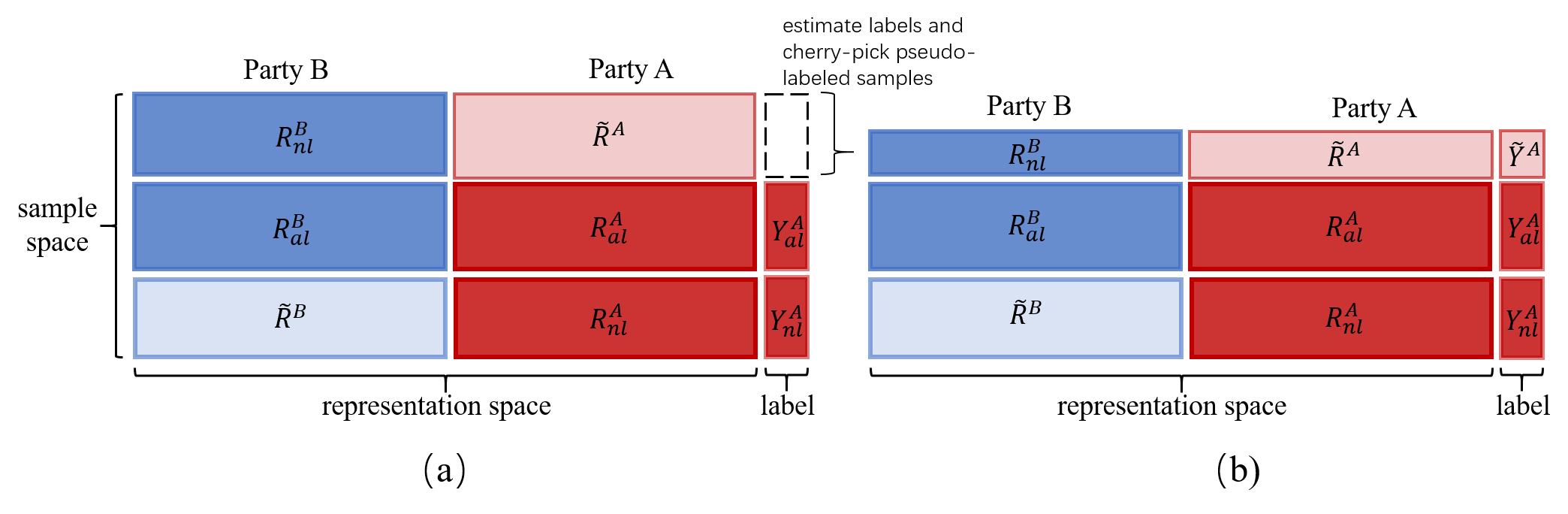}
\caption{View of the training set $\chi$, each sample of which consists of the representation of a raw input feature and its corresponding label. (a) The view of $\chi$ with missing representations estimated. (b) The view of $\chi$ with cherry-picked pseudo-labeled samples.} \label{fig_repr_tbl}
\end{figure}

We obtain estimated representations $\tilde{R}^{A}=[\tilde{R}^{A}_c;\tilde{R}^{A}_u]$ corresponds to the non-aligned representation $R^{B}_{nl}$ of party B. 
Following the same logic, we estimate the missing representations $\tilde{R}^{B}=[\tilde{R}^{B}_c;\tilde{R}^{B}_u]$ corresponds to non-aligned representation $R^{A}_{nl}$ of party A, where $\tilde{R}^{B}_{c} = g(R^{A}_{nl,c}, R^{B}_c, R^{B}_c)) $ is \emph{shared} representations and $\tilde{R}^{B}_{u} = g(R^{A}_{nl,u}, R^{A}_{al,u}, R^{B}_{al,u})$ is \emph{unique} representations. As a result, we obtain a expanded training set $\chi$ with all missing representations estimated, as depicted in Figure \ref{fig_repr_tbl}(a). However, samples $\{R^{B}_{nl},\tilde{R}^{A}\}$ in $\chi$ are unlabeled. In section \ref{est_lbl}, we will elaborate on the procedure of estimating pseudo-labels for these unlabeled samples.

To force the estimated representations to approximate the representations learned from input raw features, we add following two loss terms:

\begin{equation}\label{loss_repr}
L^{A}_{\text{diff}}(\tilde{R}^{A}_{al}, R^{A}_{al}) = \frac{1}{n_{al}}\sum_{i}\left\| \tilde{R}^{A,i}_{al}- R^{A,i}_{al}\right\|^2_F
\end{equation}

\begin{equation}
L^{B}_{\text{diff}}(\tilde{R}^{B}_{al}, R^{B}_{al}) = \frac{1}{n_{al}}\sum_{i}\left\| \tilde{R}^{B,i}_{al}- R^{B,i}_{al}\right\|^2_F, 
\end{equation}

where $\tilde{R}^{p}_{al}$ is the estimated representations of aligned samples in party $p \in \{A, B\}$ and it should be as close to $R^{p}_{al}$ as possible.

\subsection{Pseudo-Label Prediction} \label{est_lbl}

As discussed in section \ref{est_repr_sec}, samples $\{R^{B}_{nl},\tilde{R}^{A}\}$ in  training set $\chi$ are unlabeled. For each unlabeled sample indexed by $i$, we apply the three trained softmax classifiers $f^A$, $f^B$ and $f^{AB}$ that take as input $\tilde{R}^{A,i}$, $R^{B,i}_{nl}$ and $\{R^{B,i}_{nl},\tilde{R}^{A,i}\}$ respectively to produce three candidate pseudo-labels (Algorithm \ref{fedmvt_algo}, line 8). Only when all the three candidate pseudo-labels are equal and their probabilities are higher than a predefined threshold $t$ do we retain sample $i$ with the estimated pseudo-label (Algorithm \ref{fedmvt_algo}, line 9). Otherwise, we discard sample $i$ for the current iteration of training. 

As a result, all samples in the training set $\chi$ have labels, as depicted in Figure \ref{fig_repr_tbl}(b). We further define the training set $\chi=\{R^B, R^A, Y^A\}$, where $R^A$ denotes all samples of party A including estimated ones, $Y^A$ denotes labels of $R^A$, and $R^B$ denotes samples of party B that correspond to $R^A$.


\subsection{Cross-View Training} \label{mvt}

Inspired by cross-view learning \cite{clark-etal-2018-semi} to share representations across models and improve models' representation learning, we train three softmax classifiers $f^A$, $f^B$ and $f^{AB}$ jointly and each classifier takes as input a different view of the training dateset $\chi$ (Figure \ref{fig_repr_tbl}(b)) and outputs estimated class distributions. 

We define the input to classifiers $f^A$, $f^B$ and $f^{AB}$ as $\chi^A=\{R^A, Y^A\}$, $\chi^B=\{R^B, Y^A\}$, and $\chi=\{R^B, R^A, Y^A\}$, respectively. Then, the loss functions for the three classifiers are defined as follows:

\begin{equation}\label{A_loss}
L_{ce}^{A}(\chi^A) = \frac{1}{n}\sum_{i}L_{ce}(f^A(R^{A,i}), Y^{A,i})
\end{equation}
\begin{equation}\label{B_loss}
L_{ce}^{B}(\chi^B) = \frac{1}{n}\sum_{i}L_{ce}( f^B(R^{B,i}), Y^{A,i})
\end{equation}
\begin{equation}\label{AB_loss}
L_{ce}^{fed}(\chi) = \frac{1}{n}\sum_{i}L_{ce}( f^{AB}(\{R^{B,i},R^{A,i}\}), Y^{A,i})
\end{equation}

where $n$ is the number of samples in $\chi$. Formulas (\ref{A_loss}), (\ref{B_loss}) and (\ref{AB_loss}) compute the cross-entropy between estimated class distributions and ground-truth labels. Finally, we have the overall loss defined as Formula (\ref{fed_loss}):
\begin{equation}\label{fed_loss}
L_{obj} = L_{ce}^{fed} + L_{ce}^A + L_{ce}^B + \lambda_1 L^{AB}_{\text{diff}} + \lambda_2 L^{A}_{\text{diff}}+\lambda_3 L^{B}_{\text{diff}}+\lambda_4 L^A_{\text{sim}} + \lambda_5 L^B_{\text{sim}}
\end{equation}
where $\lambda_s$ are loss weights. Algorithm \ref{fedmvt_algo} gives a full version of the FedCVT algorithm.


\begin{algorithm}[h]
\caption{FedCVT algorithm}
\begin{algorithmic}[1]\label{fedmvt_algo}
\STATE \textbf{Input}: \\
Datasets $D^A$ and $D^B$; \\
Batch index sets $T_{al}$, $T^A_{nl}$ and $T^B_{nl}$ of $D_{al}$, $D^A_{nl}$ and $D^B_{nl}$, respectively.\\
Neural networks $h^B_u, h^B_c, h^A_u$ and $h^A_c$; \\
Representation estimators $g$; \\
Softmax classifiers $f^A$, $f^B$ and $f^{AB}$ \\
Class probability threshold $t$ and the epoch number $K$
\FOR{$e=1,2,...,K$}
\FOR{$b_1,b_2,b_3$ in $T_{al}, T^A_{nl}, T^B_{nl}$}
    \STATE Select mini-batches: \\
    \hspace{0.4cm} $\{X^{B,b_1}_{al}, X^{A,b_1}_{al},Y^{A,b_1}_{al}\}$ from $D_{al}$;\\
    \hspace{0.4cm} $\{X^{A,b_2}_{nl},Y^{A,b_2}_{nl}\}$ from $D^A_{nl}$;\\
    \hspace{0.4cm} $\{X^{B,b_3}_{nl}\}$ from $D^B_{nl}$;
    \STATE Learn $R^{B,b_1}_{al}, R^{B,b_3}_{nl}, R^{A,b_1}_{al}, R^{A,b_2}_{nl}$ from $X^{B,b_1}_{al}, X^{B,b_3}_{nl}, X^{A,b_1}_{al}, X^{A,b_2}_{nl}$ through $h^B_u, h^B_c, h^A_u, h^A_c$;
    \STATE Estimate missing representations $\tilde{R}^{B,b_2}$ and $\tilde{R}^{A,b_3}$ through $g$;
    \STATE Form training set $\chi$ of samples consisting of representations and corresponding labels;
    \STATE Estimate candidate pseudo-labels for unlabeled samples $\{R^{B, {b_3}}_{nl},\tilde{R}^{A, {b_3}}\}$ of $\chi$:\\
    \hspace{0.4cm} $\tilde{Y}^{b_3}_1 = f^A(\tilde{R}^{A, {b_3}})$;\\
    \hspace{0.4cm} $\tilde{Y}^{b_3}_2 = f^B(R^{B, {b_3}}_{nl})$;\\
    \hspace{0.4cm} $\tilde{Y}^{b_3}_3 = f^{AB}([R^{B, {b_3}}_{nl};\tilde{R}^{A, {b_3}}]);$
    \STATE Retain samples of $\{R^{B, {b_3}}_{nl},\tilde{R}^{A, {b_3}}, (\tilde{Y}^{b_3}_1,\tilde{Y}^{b_3}_2,\tilde{Y}^{b_3}_{3})\}$ that satisfy the rule defined in section \ref{est_lbl};
    \STATE Feed $\chi^A$, $\chi^B$ and $\chi$ to classifiers $f^A$, $f^B$ and $f^{AB}$ respectively for cross-view training;
    \STATE Compute loss: $L_{obj} = L^{fed} + L^A + L^B + \lambda_1 L^{AB}_{\text{diff}} + \lambda_2 L^{A}_{\text{diff}}+\lambda_3 L^{B}_{\text{diff}}+\lambda_4 L^A_{\text{sim}} + \lambda_5 L^B_{\text{sim}}$;
    \STATE Compute gradients $\nabla L_{obj}$ and update models.
\ENDFOR
\ENDFOR
\end{algorithmic}
\end{algorithm}

\subsection{Security Analysis} \label{privacy}

FedCVT does not require participating parties to share their original data and models (including architecture and parameters) but only intermediate representations and gradients. The intermediate representations are results transformed from original features through deep neural networks with multiple layers of non-linear transformations. There is little chance for a malicious party to reconstruct the victim party's original features from exposed intermediate representations without knowing the victim's model \cite{splitNN@Otkrist}. In recent years, there has been a series of works on investigating data leakage through gradients~\cite{zhu2020deep,zhao2020idlg,geiping2020inverting,gradinv2021}. However, those research works focus on horizontal federated learning where the server is malicious. They assume that the malicious server can access the victim client's model parameters and gradients. These assumptions are not held in the FedCVT setting. Research works~\cite{liuyang2020backdoor,li2021label} investigate privacy leakage issues in the VFL setting and demonstrate that the malicious party could recover party A's labels through gradients. To provide a strong privacy guarantee, FATE design and implement a privacy-preserving VFL-DNN (Deep Neural Network) framework that can efficiently perform encryption in both forward and backward stages of the VFL training procedure to prevent intermediate representations and gradients from being exposed. For experimental convenience, we did not implement FedCVT with FATE. However, FedCVT is compatible with the VFL-DNN framework and can be migrated to FATE straightforwardly.

\section{Experimental Evaluation}

We report experiments conducted on three public datasets, which are: 1) NUS-WIDE dataset \cite{Chua09nus-wide:a}; 2) Vehicle dataset~\cite{DUARTE2004826}; 3) CIFAR-10  \cite{Krizhevsky09learningmultiple} to validate our proposed FedCVT approach.


\subsection{Hyperparameters}

FedCVT-VFL combines multiple techniques to improve the performance of VFL and thus it introduces various hyperparameters - specifically, the loss weights  $\lambda_s$, the label probability threshold $t$, the sharpening temperature $T$ for estimating representation and learning rate. For experiments on NUS-WIDE, we set $\lambda_{1,2,3,4,5}=0.1$, $t$=0.5, $T$=0.5 and learning rate = 0.005. For Vehicle, we set $\lambda_{1,2,3,4,5}=0.1$, $t$=0.5, $T$=1.0 and learning rate = 0.005. For CIFAR-10, we set $\lambda_{1,4,5}=0.1$, $\lambda_{2,3}=0.01$, $t$=0.6, $T$=0.1 and learning rate = 0.001. For all experiments, we use Adam \cite{adam_KingmaB14} optimizer.

\subsection{Baseline Models}
We consider three baseline models: (1) party A's local model $f^{A}$, (2) vanilla-VFL and (3) FTL~\cite{SecureFTL}. Party A's local model is trained on \textit{all local samples} of party A, while vanilla-VFL is trained on \textit{aligned samples} between party A and party B (without using non-aligned samples). FTL leverages all samples of party A in addition to aligned samples to help party B build a classifier model. We consider party A's local model as a baseline model because it is worth examining whether VFL performs better than the local model when the number of aligned samples is limited. Vanilla-VFL and FTL are adopted to verify whether our FedCVT-VFL approach is effective. Note that FTL aims to build a model only for party B, and thus it does not have a federated model that takes input from both parties. To ensure a fair comparison, we add a federated model on top of local models of FTL and denote this modified FTL as mFTL.

\subsection{Experimental Results}

\subsubsection{NUS-WIDE} The NUS-WIDE dataset consists of 634 low-level image features extracted from Flickr images and their associated 1000 textual tags as well as 81 ground truth labels. In this work, we consider solving a 10-label classification problem with a data federation formed between party A and party B. In our setting, each party has 56000 training samples, 8000 aligned validation samples, and 10000 aligned testing samples. Each party utilizes two local neural networks that each has one hidden layer with 96 units to learn representations from local input samples. Then, each party feeds the learned representations into its local softmax classifier $f^{p}, p \in \{A,B\}$ with the dimension of 192 (96x2) and the federated softmax layer $f^{AB}$ with the dimension of 384 (96x2x2), respectively, for federated training.

On NUS-WIDE, we run experiments with two scenarios. In the first scenario, denoted as scenario-1, we put 634 image features on party A and 1000 textual features on party B. The second scenario, denoted as scenario-2, is the other way around. In both scenarios, party A owns the labels. We consider these two scenarios because image and text are different in modality and have different discriminative power, and we want to investigate how these differences would affect the performance of FedCVT-VFL.

\begin{table}[h]
\centering
\begin{tabular}{m{5.5em}| m{4.5em} m{4.5em} m{4.5em} m{4.5em} m{4.5em} m{4.5em}}
\hline
\multicolumn{7}{c}{scenario-1: Party A with image} \\\hline
 \multicolumn{1}{c}{Model} & \multicolumn{1}{c}{250} & \multicolumn{1}{c}{500} &\multicolumn{1}{c}{1000} & \multicolumn{1}{c}{2000} & \multicolumn{1}{c}{4000} & \multicolumn{1}{c}{8000} \\
\hline
\hline
{\small Local Model} & $58.92$ & $58.92$ & $58.92$ & $58.92$ & $58.92$ & $58.92$ \\
{\small Vanilla-VFL} & $51.79\pm0.21$ & $56.58\pm0.28$ & $64.23\pm0.09$ & $68.05\pm0.36$ & $71.07\pm0.14$ & $73.69\pm0.12$ \\
{\small mFTL} &$62.43\pm0.46$ & $65.07\pm0.29$ & $68.49\pm0.36$ & $71.19\pm0.26$ & $ 73.10\pm0.27$ & $74.32\pm0.36$ \\
{\small FedCVT-VFL} & $65.16\pm0.21$ & $68.37\pm0.22$ & $68.93\pm0.20$ & $71.75\pm0.14$ & $74.27\pm0.24$ & $75.45\pm0.11$ \\
$\Delta$ \footnotesize{FedCVT-VFL} & $\uparrow 13.37$ & $\uparrow 11.79$ & $\uparrow 4.7$ & $\uparrow 3.7$ & $\uparrow 3.2$ & $\uparrow 1.76$ \\
\hline
\multicolumn{7}{c}{scenario-2: Party A with text} \\\hline
 \multicolumn{1}{c}{Model} & \multicolumn{1}{c}{250} & \multicolumn{1}{c}{500} &\multicolumn{1}{c}{1000} & \multicolumn{1}{c}{2000} & \multicolumn{1}{c}{4000} & \multicolumn{1}{c}{8000} \\
\hline
{\small Local Model} & $70.31$ & $70.31$ & $70.31$ & $70.31$ & $70.31$ & $70.31$ \\
{\small Vanilla-VFL} & $51.79\pm0.21$ & $56.58\pm0.28$ & $64.23\pm0.09$ & $68.05\pm0.36$ & $71.07\pm0.14$ & $73.69\pm0.12$ \\
{\small mFTL} &$63.95\pm0.26$ & $67.17\pm0.28$ & $70.34\pm0.19$ & $72.59\pm0.20$ & $ 73.79\pm0.18$ & $74.51\pm0.18$ \\
{\small FedCVT-VFL} &$68.76\pm0.08$ & $71.51\pm0.27$ & $72.77\pm0.21$ & $73.95\pm0.06$ & $ 75.92\pm0.24$ & $77.45\pm0.06$ \\
$\Delta$ \footnotesize{FedCVT-VFL} & $\uparrow 16.97$ & $\uparrow 14.93$ & $\uparrow 8.54$ & $\uparrow 5.9$ & $\uparrow 4.85$ & $\uparrow 3.76$ \\
\hline
\end{tabular}
    \caption{Test accuracy (\%) comparison of FedCVT-VFL to party A's local model, mFTL and vanilla-VFL on NUS-WIDE for a varying number of aligned samples. Note that the local model is trained based on 56000 local samples of party A and thus it is not affected by the number of aligned samples. $\Delta$ is the performance gain of FedCVT-VFL compared to Vanilla-VFL.}
    \label{nus_wide_table}
\end{table}

\begin{figure}[h]
\centering
    \subfigure[]{\includegraphics[width=0.48\linewidth,height=4.5cm]{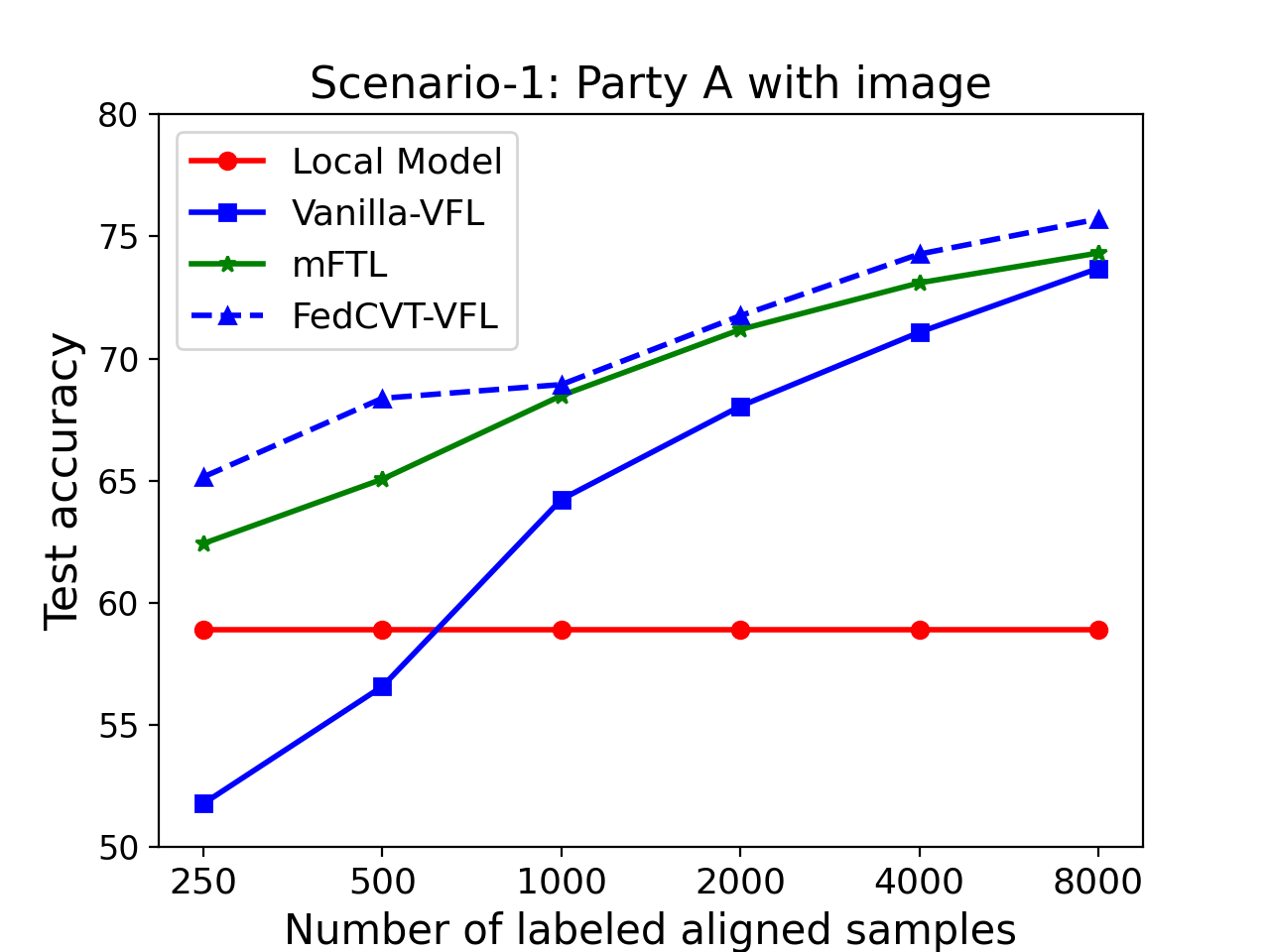} \label{party_a_image}}
    \subfigure[]{\includegraphics[width=0.48\linewidth,height=4.5cm]{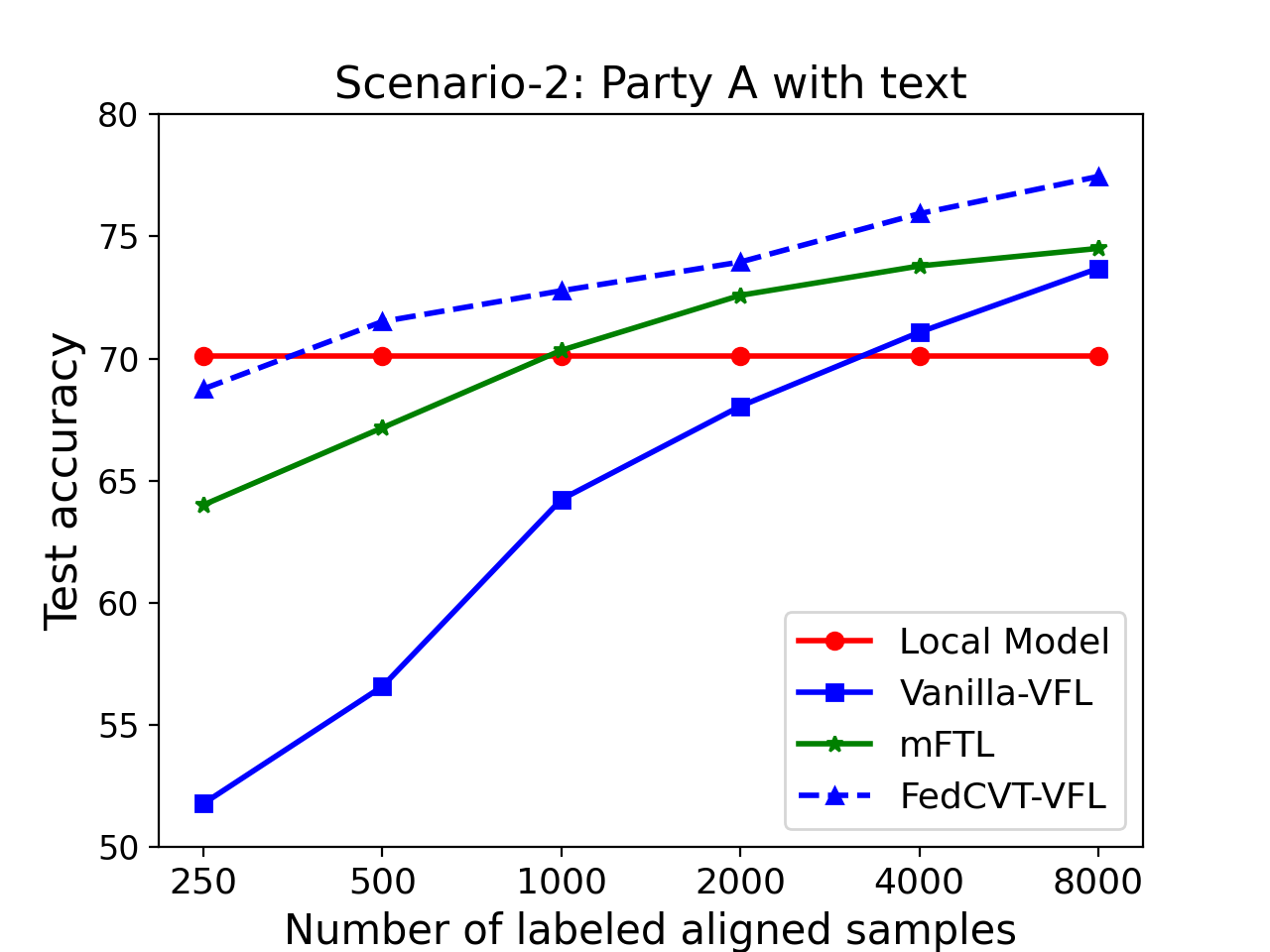} \label{party_a_text}} 
    \caption{Test accuracy (\%) comparison of FedCVT-VFL model to vanilla-VFL model on NUS-WIDE for a varying number of aligned samples. Exact numbers are provided in Table \ref{nus_wide_table}. (a) shows results when party A holds image data, while (b) shows results when party A hold text data.}
    \label{nus_wide_fig}
\end{figure}

Table \ref{nus_wide_table} shows the accuracy comparison between FedCVT-VFL and baselines with a varying number of aligned samples from 250 to 8000. FedCVT-VFL outperforms mFTL~\cite{SecureFTL} noticeably and outperforms vanilla-VFL by a large margin (see Figure 6) for both scenarios. For example, with 250 aligned samples, FedCVT-VFL improves 13.4 and 17.0 points in accuracy over vanilla-VFL in scenario-1 and scenario-2, respectively. With 8000 aligned samples, FedCVT-VFL enhances the accuracy by 1.8 and 3.8 points, respectively. In addition, FedCVT-VFL outperforms the local model with only a small amount of aligned samples. For example, FedCVT-VFL outperforms the local model with only 250 aligned samples in scenario-1 and with 500 aligned samples in scenario-2.  

As shown in Table \ref{nus_wide_table}, the local model of scenario-2 significantly outperforms that of scenario-1 (by 11.4 points), which demonstrates that text features have much more discriminative power than image features on the NUS-WIDE dataset. Therefore, in scenario-2, leveraging complementary image features contributes little to vanilla-VFL when the number of aligned samples is small. This partially explains why vanilla-VFL models do not outperform the local model (solid-blue line v.s. red line in Figure \ref{party_a_text}) in scenario-2 when the size of aligned samples is not sufficiently large enough (i.e., $\leq 2000$). While in scenario-1, party A benefits significantly from leveraging complementary text features of party B through VFL (solid-blue line v.s. red line in Figure \ref{party_a_image}). This suggests that, in practice, when the size of the aligned samples is limited, the party owning labels may be better off using its local model (with much more training samples) if it has much stronger discriminative features than other parties. Otherwise, it may benefit from leveraging the features of other parties.

\subsubsection{Vehicle} The vehicle\footnote{\url{https://www.csie.ntu.edu.tw/~cjlin/libsvmtools/datasets/multiclass.html}} \cite{DUARTE2004826} dataset is for classifying the types of moving vehicles in a distributed sensor network. There are 3 types of vehicles. Each sample in this dataset has 100 features, the first 50 of which are $\textit{acoustic}$ features, while the rest are $\textit{seismic}$ features. Thus, it is natural to split the two modalities of features into two parties. In our setting, each party has 40000 training samples, 5000 validation samples, and 5000 testing samples. Each party utilizes two local neural networks that each has one hidden layer with 32 units to learn representations from local input samples. Then, each party feeds the learned representations into its local softmax classifier $f^{p}, p \in \{A,B\}$ with the dimension of 64 (32x2) and the federated softmax layer $f^{AB}$ with the dimension of 128 (32x2x2), respectively, for federated training.

Similar to NUS-WIDE, we consider two scenarios on the Vehicle dataset. In scenario-1, we put acoustic features on party A. While in scenario-2, we put seismic features on party A. In both scenarios, party A owns the labels. 

\begin{figure}[h]
\centering
    \subfigure[]{\includegraphics[width=0.48\linewidth,height=4.5cm]{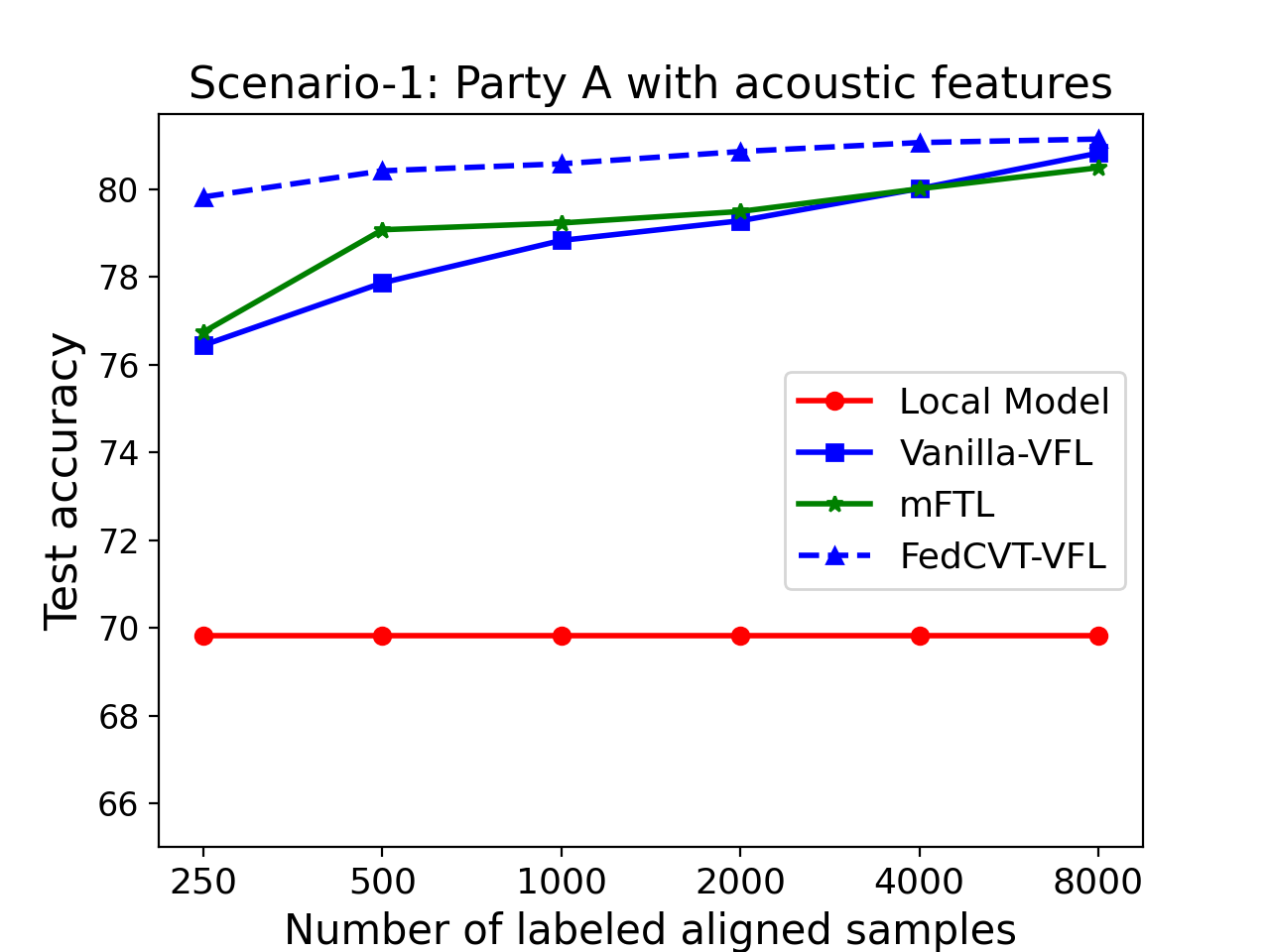} \label{party_a_acoustic}}
    \subfigure[]{\includegraphics[width=0.48\linewidth,height=4.5cm]{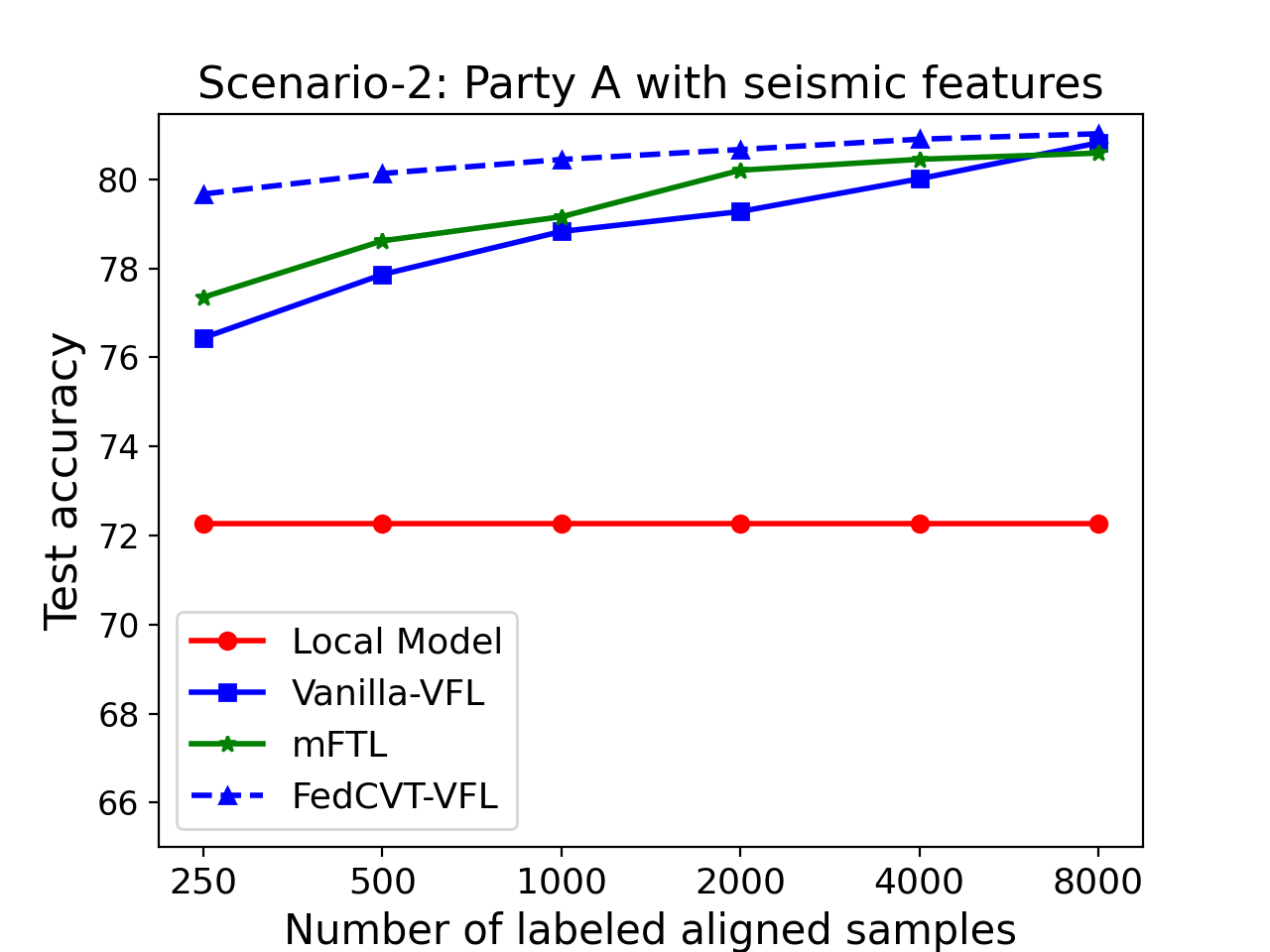} \label{party_a_seismic}} 
    \caption{Test accuracy (\%) comparison of FedCVT-VFL model to vanilla-VFL on Vehicle dataset for a varying number of aligned samples. Exact numbers are provided in Table \ref{vehicle_table}. (a) shows results when party A holds acoustic data, while (b) shows results when party A hold seismic data.}
    \label{vehicle_fig}
\end{figure}

\begin{table}[h]
\centering
\begin{tabular}{m{5.5em}| m{4.5em} m{4.5em} m{4.5em} m{4.5em} m{4.5em} m{4.5em}}
\hline
\multicolumn{7}{c}{scenario-1: Party A with acoustic features} \\\hline
 \multicolumn{1}{c}{Model} & \multicolumn{1}{c}{250} & \multicolumn{1}{c}{500} &\multicolumn{1}{c}{1000} & \multicolumn{1}{c}{2000} & \multicolumn{1}{c}{4000} & \multicolumn{1}{c}{8000} \\
\hline
\hline
{\small Local Model} & $69.91$ & $69.91$ & $69.91$ & $69.91$ & $69.91$ & $69.91$ \\
{\small Vanilla-VFL} & $76.44\pm0.64$ & $77.86\pm0.83$ & $78.83\pm0.53$ & $79.28\pm0.75$ & $80.24\pm0.60$ & $80.82\pm0.29$ \\
{\small mFTL} &$76.74\pm0.19$ & $79.07\pm0.17$ & $79.22\pm0.18$ & $79.49\pm0.23$ & $ 80.01\pm0.21$ & $80.49\pm0.20$ \\
{\small FedCVT-VFL} & $79.82\pm0.27$ & $80.42\pm0.28$ & $80.58\pm0.07$ & $80.86\pm0.09$ & $81.05\pm0.17$ & $81.14\pm0.06$ \\
$\Delta$ \footnotesize{FedCVT-VFL} & $\uparrow 3.38$ & $\uparrow 2.56$ & $\uparrow 1.75$ & $\uparrow 1.58$ & $\uparrow 0.81$ & $\uparrow 0.32$ \\
\hline
\multicolumn{7}{c}{scenario-2: Party A with seismic features} \\\hline
 \multicolumn{1}{c}{Model} & \multicolumn{1}{c}{250} & \multicolumn{1}{c}{500} &\multicolumn{1}{c}{1000} & \multicolumn{1}{c}{2000} & \multicolumn{1}{c}{4000} & \multicolumn{1}{c}{8000} \\
\hline
{\small Local Model} & $70.31$ & $70.31$ & $70.31$ & $70.31$ & $70.31$ & $70.31$ \\
{\small Vanilla-VFL} & $76.44\pm0.64$ & $77.86\pm0.83$ & $78.83\pm0.53$ & $79.28\pm0.75$ & $80.24\pm0.60$ & $80.82\pm0.29$ \\
{\small mFTL} &$77.35\pm0.39$ & $78.62\pm0.35$ & $79.16\pm0.18$ & $80.21\pm0.19$ & $ 80.45\pm0.23$ & $80.59\pm0.17$ \\
{\small FedCVT-VFL} &$79.67\pm0.35$ & $80.13\pm0.25$ & $80.44\pm0.16$ & $80.67\pm0.16$ & $ 80.90\pm0.15$ & $81.02\pm0.05$ \\
$\Delta$ \footnotesize{FedCVT-VFL} & $\uparrow 3.23$ & $\uparrow 2.27$ & $\uparrow 1.61$ & $\uparrow 1.39$ & $\uparrow 0.66$ & $\uparrow 0.20$ \\
\hline
\end{tabular}
    \caption{Test accuracy (\%) comparison of FedCVT-VFL to party A's local model, mFTL and vanilla-VFL model on Vehicle dataset for a varying number of aligned samples. Note that the local model is trained based on 40000 local samples of party A and thus it is not affected by the number of aligned samples. $\Delta$ is the performance gain of FedCVT-VFL compared to Vanilla-VFL.}
    \label{vehicle_table}
\end{table}

From Table \ref{vehicle_table} and Figure \ref{vehicle_fig}, we have the following observations. For both scenarios, FedCVT and vanilla-VFL outperform the local model by a large margin, even with only 250 aligned samples. This demonstrates that leveraging more complementary features brings great benefits for improving the accuracy of machine learning models on the Vehicle dataset. Compared to vanilla-VFL and mFTL, FedCVT achieves better performance with much less aligned samples. For example, in both scenarios, with 500 aligned samples, FedCVT reaches an accuracy comparable to the performance of vanilla-VFL and mFTL using 4000 aligned samples. It is worth noting from Figure \ref{vehicle_fig} that mFTL becomes less accurate than vanilla-VFL when the number of aligned samples reaches 8000. We conjecture that this is because mFTL mainly leverages two parties' shared representations that are learned based on aligned samples to boost performance, and thus it lost information unique to each party. With more aligned samples are added for training, more unique information would be lost.


\subsubsection{CIFAR-10.} The CIFAR-10 dataset consists of color images with shape of $32 \times 32 \times 3$ in 10 classes. To simulate the two-party vertical federated learning setting, we partition each CIFAR-10 image with shape $32 \times 32 \times 3$ vertically into two parts (each part has shape $32 \times 16 \times 3$). Each party has 25000 training samples, 5000 aligned validation samples, and 10000 aligned testing samples. Each party uses two local VGG-like CNN models to learn representations from input image samples. Then it feeds the learned representations into its local softmax classifier $f^{p}, p \in \{A,B\}$ and the federated softmax classifier $f^{AB}$, respectively, for jointly training. The VGG-like CNN model consists of 2x2 max-pooling layers, 3x3 convolutional layers with stride 1 and padding 1, and fully connected layers. The architecture is \textbf{conv32-conv32-pool-conv64-conv64-pool-conv128-conv128-pool-fc64}. Thus, the federated softmax classifier has the dimension of 256 (64x2x2). For CIFAR10, we do not swap image partitions between the two parties because they are cut from the same images, and thus we assume they have similar discriminative power. 

\begin{figure}[h]
\centering
    \includegraphics[width=0.50\linewidth,height=4.5cm]{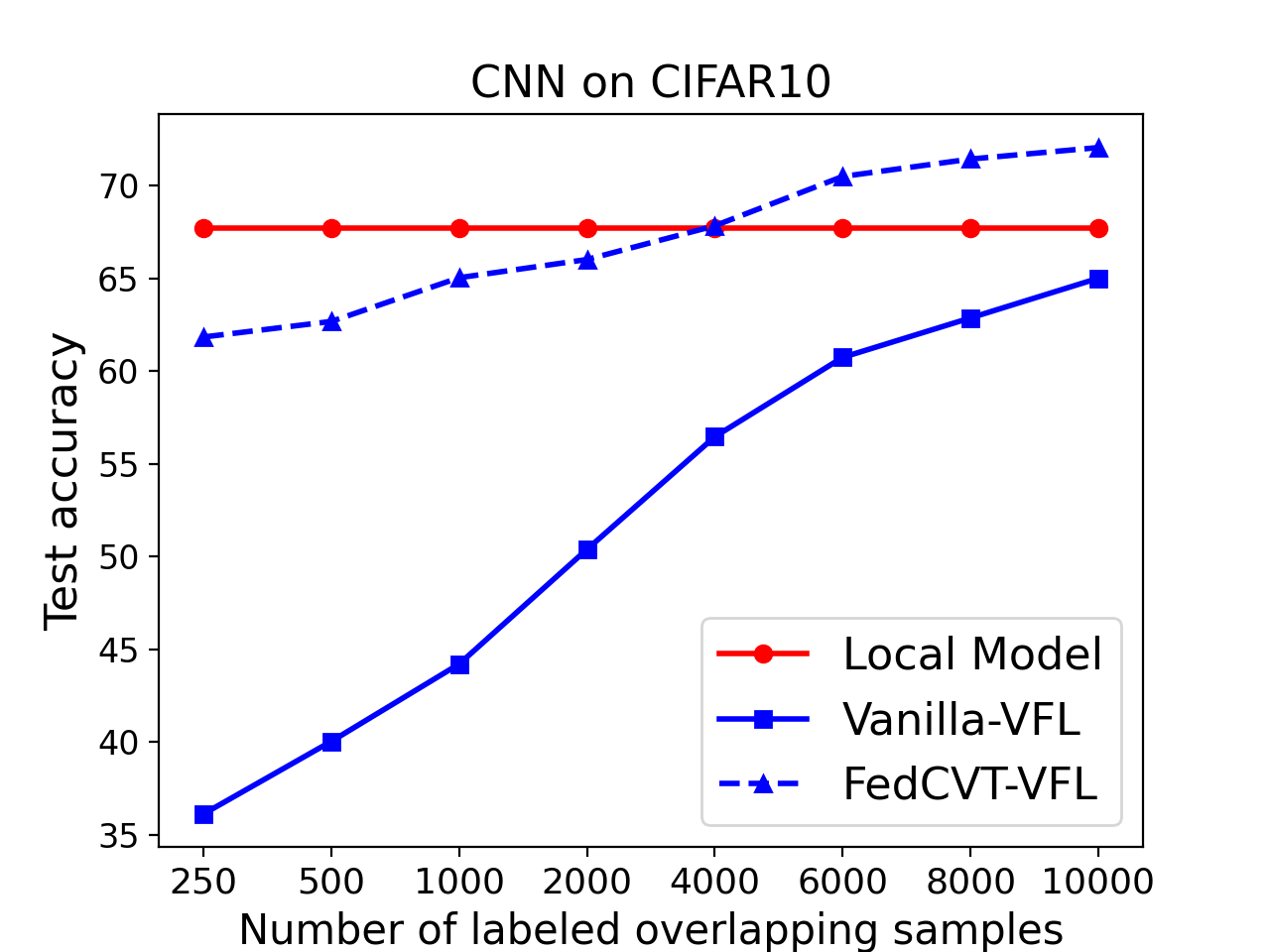}
    \caption{Test accuracy (\%) comparison of FedCVT-VFL model to vanilla-VFL model on CIFAR10 for a varying number of aligned samples. Exact numbers are provided in Table \ref{cifar10_table}.}
    \label{cifar10_fig}
\end{figure}

\begin{table}[h]
\centering
\begin{tabular}{m{5.5em}| m{4.4em} m{4.4em} m{4.4em} m{4.4em} m{4.4em} m{4.5em}}
\hline
 \multicolumn{1}{c}{Model} & \multicolumn{1}{c}{500} & \multicolumn{1}{c}{1000} &\multicolumn{1}{c}{2000} & \multicolumn{1}{c}{4000} & \multicolumn{1}{c}{8000} & \multicolumn{1}{c}{10000} \\
\hline
\hline
{\small Local Model} & $67.74$ & $67.74$ & $67.74$ & $67.74$ & $67.74$ & $67.74$ \\
{\small Vanilla-VFL} & $40.02\pm0.13$ & $44.20\pm0.12$ & $50.38\pm0.14$ & $56.46\pm0.06$ & $62.87\pm0.08$ & $65.01\pm0.11$\\
{\small FedCVT-VFL} & $62.68\pm0.21$ & $65.04\pm0.26$ & $66.02\pm0.31$ & $67.84\pm0.23$ & $71.44\pm0.11$ & $72.06\pm0.16$ \\
$\Delta$ \footnotesize{FedCVT-VFL} & $\uparrow 22.66$ & $\uparrow 20.84$ & $\uparrow 15.64$ & $\uparrow 11.38$ & $\uparrow 8.57$ & $\uparrow 7.05$ \\
\hline
\end{tabular}
    \caption{Test accuracy (\%) comparison of FedCVT-VFL model to vanilla-VFL model on CIFAR-10 for a varying number of aligned samples. Note that the local model is trained based on 25000 local samples of party A and thus it is not affected by the number of aligned samples. $\Delta$ is the performance gain of FedCVT-VFL compared to Vanilla-VFL.}
    \label{cifar10_table}
\end{table}

As shown in Table \ref{cifar10_table}, FedCVT-VFL significantly outperforms vanilla-VFL for CIFAR10 (dashed-blue line v.s. solid-blue line in Figure \ref{cifar10_fig}). For example, FedCVT-VFL improves accuracy by 22.7 points with 500 aligned samples and 7.1 points with 10000 aligned samples compared with vanilla-VFL. 
Figure \ref{cifar10_fig} (solid-red line v.s. solid-blue line) depicts that the vanilla-VFL cannot beat the local model with even 10000 aligned samples (over 1/3 of local samples). FedCVT-VFL boosts the performance of vanilla-VFL so that it outperforms the local model using 4000 aligned samples (less than 1/6 of local samples). This manifests that FedCVT equips VFL with an effective way of leveraging unlabeled data to improving model performance.

\subsection{Ablation Study}

FedCVT exploits several techniques to improve performance. It leverages unique and shared representations learned from raw features, uses both types of representations to estimate missing representations, and then performs cross-view training with additional pseudo-labeled samples. In this section, we study how each of those techniques contributes to the performance of FedCVT. Specifically, we measure effects of:

\begin{itemize}
    \item Use both unique and shared representations without representation estimation and cross-view training.
    \item Use only unique representations, only shared representations, or both unique and shared representations for estimating missing representations but without cross-view training.
    \item All techniques applied.
\end{itemize}

\begin{table}[h]
\centering
\begin{tabular}{m{28em} m{2.1em} m{2.1em}}
\hline
 \multicolumn{1}{c}{Ablation} & \multicolumn{1}{c}{500} & \multicolumn{1}{c}{4000} \\
\hline
\hline
{\small (1) Vanilla VFL (equals to using only unique representations w/o RE and CVT)} & 56.58 & 71.07 \\
{\small (2) FedCVT-VFL uses unique and shared representations w/o RE and CVT} & 63.10 & 72.70 \\
{\small (3) FedCVT-VFL uses only unique representations with RE w/o CVT} & 66.12 & 73.31 \\
{\small (4) FedCVT-VFL uses only shared representations with RE w/o CVT} & 65.68 & 72.89 \\
{\small (5) FedCVT-VFL uses unique and shared representations with RE w/o CVT} & 67.66 & 73.69 \\
{\small (6) FedCVT-VFL with all techniques applied} & 68.37 & 74.27  \\
\hline
\end{tabular}
    \caption{Ablation study results. All values are accuracy (\%) on NUS-WIDE (scenario 1) with 500 or 4000 aligned samples.}
    \label{ablation}
\end{table}


We conducted the ablation study on NUS-WIDE (scenario-1) with 500 and 4000 aligned samples. For conciseness, we denote representation estimation as RE and cross-view training as CVT. The results are reported in Table \ref{ablation}. It shows that every technique applied to FedCVT-VFL contributes to the performance boost. We obtain the most dramatic performance gain when applying representation estimation. For example, comparing (5) and (1), FedCVT-VFL outperforms vanilla-VFL by 11\% and 2.6\%, respectively, using 500 samples and 4000 samples. Comparing (6) and (5), cross-view training also helps boost the performance of FedCVT-VFL.

\section{Conclusion}
We propose Federated Cross-View Training (FedCVT), a semi-supervised learning approach that improves the performance of VFL using limited aligned samples. FedCVT leverages representation estimation and pseudo-labels prediction to expand the training set and trains three classifiers jointly through vertical federated learning to improve models' representation learning. FedCVT-VFL significantly enhances the performance of vanilla VFL. The ablation study shows that each technique used in FedCVT-VFL contributes to the final performance boost.

\section{Acknowledgment}
This work was partially supported by the National Key Research and Development Program of China under Grant No. 2018AAA0101100.

\bibliographystyle{ACM-Reference-Format}
\bibliography{ref}

\end{document}